\documentclass[conference]{IEEEtran}
\IEEEoverridecommandlockouts
\usepackage{cite}
\usepackage{amsmath,amssymb,amsfonts}
\usepackage{algorithmic}
\usepackage{graphicx}
\usepackage{stfloats}
\usepackage{textcomp}
\usepackage{xcolor}
\def\BibTeX{{\rm B\kern-.05em{\sc i\kern-.025em b}\kern-.08em
    T\kern-.1667em\lower.7ex\hbox{E}\kern-.125emX}}

\begin{document}
\title{{Text-image guided Diffusion Model for generating Deepfake celebrity interactions}
}

\author{

\IEEEauthorblockN{Yunzhuo Chen}
\IEEEauthorblockA{\textit{The University of Western Australia} \\
Perth, Australia \\
yunzhuo.chen@research.uwa.edu.au}
\and

\IEEEauthorblockN{Nur Al Hasan Haldar}
\IEEEauthorblockA{\textit{The University of Western Australia} \\
Perth, Australia \\
nur.haldar@uwa.edu.au}

\IEEEauthorblockN{Ajmal Mian}
\IEEEauthorblockA{\textit{The University of Western Australia} \\
Perth, Australia \\
ajmal.mian@uwa.edu.au} 
\and

\IEEEauthorblockN{Naveed Akhtar}
\IEEEauthorblockA{\textit{The University of Western Australia} \\
Perth, Australia \\
naveed.akhtar@uwa.edu.au}
\\

}

\maketitle

\begin{abstract}
Deepfake images are fast becoming a serious concern due to their realism.  Diffusion models have recently demonstrated highly realistic visual content generation, which makes them an excellent potential tool for Deepfake generation. To curb their exploitation for Deepfakes, it is imperative to first explore the extent to which diffusion models can be used to generate realistic content that is controllable with convenient prompts. This paper devises and explores a novel method in that regard. Our technique alters the popular stable diffusion model to generate a controllable high-quality Deepfake image with text and image prompts. In addition, the original stable model lacks severely in generating quality images that contain multiple persons.  The modified diffusion model is able to address this problem, it add input anchor image's latent at the beginning of inferencing rather than Gaussian random latent as input. Hence, we focus on generating forged content for celebrity interactions, which may be used to spread rumors. We also apply Dreambooth to enhance the realism of our fake images. Dreambooth trains the pairing of center words and specific features to produce more refined and personalized output images. Our results show that with the devised scheme, it is possible to create fake visual content with alarming realism, such that the content can serve as believable evidence of meetings between powerful political figures.

\end{abstract}


\section{Introduction}

Deepfake~\cite{westerlund2019emergence} face-forgery technology first appeared on the community forum Reddit~\cite{roettgers2018porn}. Since then, it has set off a wave of creating fake content~\cite{bitesizedeepfakes}. 
The advancement of Deepfake generative models~\cite{kemelmacher2016transfiguring,thies2016face2face,koujan2020head2head,nirkin2019fsgan,pumarola2018ganimation,wu2018reenactgan,natsume2018rsgan,suwajanakorn2017synthesizing}  can easily misguide human visual system.  Deepfake technology, despite its potential benefits in the field of entertainment, also raises serious concerns for its potential negative societal impact. The ease of access to software and codes for creating fake videos and images has resulted in a widespread abuse of  this technology, leading to legal violations such as creating illegal or offensive content~\cite {roettgers2018porn}.  Therefore, it is crucial to address and mitigate the negative consequences of its abuse. Currently, Deepfake content is generally still human-identifiable. This is because existing Deepfake generation processes usually focus on specific facial regions, which introduces  discernible artifacts in the overall content. However, recent advances in computer vision have seen a quantum leap in the quality of visual content generation through generative models, such as diffusion models~\cite {rombach2022high,wijmans1995solution,saharia2022palette,ho2022cascaded,gu2022vector}. This makes these models a potent potential tool for Deepfake generation.

\begin{figure}[t]
\centering
\includegraphics[ width= 0.47\textwidth]{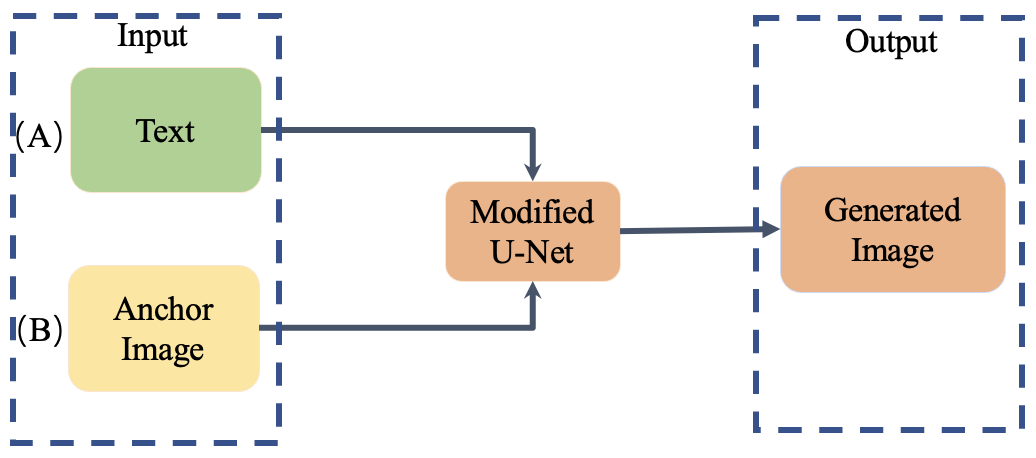}
\caption{Two-stream input U-Net consisting of (A) User input prompt; (B) User input image. The utilization of an anchor image offers a significant advantage in mitigating the uncertainty introduced by Gaussian noise. The anchor image serves as a valuable reference point, enabling the model to generate images that align closely with the desired characteristics and features depicted in the anchor image. This helps to minimize the inherent uncertainty and enhances the overall quality and accuracy of the generated content.}
\end{figure}

To curb the negative impact of Deepfake, it is imperative to thoroughly explore the capacity of diffusion models~\cite{sohl2015deep, song2019generative, koujan2020head2head, pumarola2018ganimation} as a tool to generate controlled fake content. To that end, we focus on the widely popular publicly accessible stable diffusion model~\cite {rombach2022high}, which has recently gained a strong reputation for its generative abilities.
We find that whereas stable diffusion generates high-quality synthetic content, its off-the-shelf use for controlled and topic-oriented fake content generation still requires some effort. The random seed used by the model in the generation process is the main reason for this phenomenon. The generated images suffer from uncontrollable background and disfigured foreground when off-the-shelf model is used to generate scenes with multiple characters. Especially, the forged handshake details are easily identifiable, even in the good quality generated samples. 

In this work, we explore ways to significantly improve the quality of the visual content generated by the stable diffusion model~\cite {rombach2022high} by extending the original design, such that the generated images can serve  as believable evidence for significant  events, e.g., meetings between powerful political figures. In particular, our approach is able to provide convenient text and image handles over the content which enables controlling the background and foreground content.  Our proposed Text-Image Guided Diffusion Model (TIDM) leverages an input image as guidance to improve the content quality of the generated images, which also enables generating consistent multiple images of an event, and better meet the needs of the Deepfake task. To prevent output randomness, our approach incorporates an anchor image and text into the model's input. Additionally, we introduce a novel technique, inspired by Dreambooth~\cite{ruiz2022dreambooth}, to streamline the training of the modified model, ensuring convenience and efficiency.

We conducted comprehensive experiments using four different metrics to evaluate our approach. Through meticulous experimentation and ablation studies, we demonstrate the efficacy of our approach. Our findings underscore the potential risks associated with using generative modeling to create convincing fake content as evidence of influential events. They also emphasize the importance of implementing practical measures to mitigate the adverse impacts of Deepfake on society. The contributions of this paper can be summarized as follows.

\begin{itemize}
    \item This work addresses the Deepfake generation problem from a diffusion model perspective, considering both characters and backgrounds in the Deepfake generation task as a whole.
    \item To reduce generative randomness in the diffusion model, we enhance the underlying traditional U-Net architecture by incorporating not only text but also the latent representation of an anchor image related to the generated content.
    \item By leveraging Dreambooth technique, we propose an efficient training method for our proposed model.
    \item We validate the effectiveness of our technique through extensive experiments and ablation studies.
\end{itemize}

\begin{figure*}[t]
\centering
\includegraphics[ width= 1\textwidth]{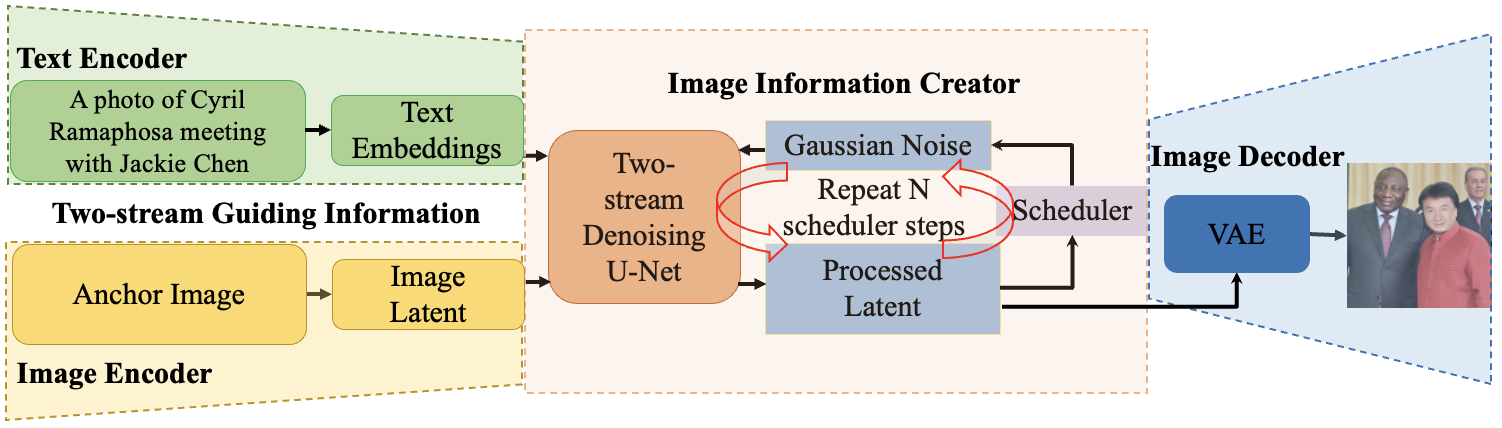}
\vspace{-4mm}
\caption{The original stable diffusion model~\cite{rombach2022high} consists of three modules: Text Encoder,  Image Information Creator, and Image Decoder. Our technique introduces an additional Image Encoder that allows leveraging an anchor image. The  proposed model uses a two-stream denoising U-Net to process the additional data modality in the Image Information Creator block. Our model  produces high-quality realistic image(s) that can depict interactive event, e.g., meetings, with consistent multiple forged images. }
\label{Inference process with new U-net architecture}
\end{figure*}

\section{RELATED WORK}
\label{sec:RW}

The rapid development and exploration of robust and powerful Deepfake generation methods are essential to address the increasing security concerns arising from forged visual content. Numerous deepfake generation techniques employing generative adversarial networks (GANs) have been extensively studied in the literature to investigate the security risks associated with generative visual modeling~\cite{kemelmacher2016transfiguring, thies2016face2face, koujan2020head2head, nirkin2019fsgan, pumarola2018ganimation, wu2018reenactgan, natsume2018rsgan, suwajanakorn2017synthesizing}. Early research in this domain predominantly focused on constructing Deepfake generative networks using neural networks. However, the advancement of generative models in the deep learning field has been comparatively slower than that of discriminative models. The introduction of Generative Adversarial Networks (GANs) by Goodfellow et al.~\cite{goodfellow2020generative} has significantly enhanced the capabilities of generative models, revolutionizing the field of generative visual modeling.

ProGAN~\cite{karras2017progressive} and StyleGAN~\cite{karras2019style} techniques have achieved remarkable success in generating high-resolution images with desired features. These methods, along with further research on novel architectures and their application, hold significant potential for advancing Deepfake generation. However, existing Deepfake generation methods still face certain limitations. One limitation involves the lack of realistic backgrounds in the generated images~\cite{wu2018reenactgan,natsume2018rsgan}. Blurring the background to focus on the face can result in distribution differences, as highlighted in studies such as~\cite{karras2019style}. Moreover, most face forgery techniques only modify specific regions of the face, which often involves operations like cropping and splicing. These operations introduce counterfeit traces and artifacts. For instance, the process of blurring the face contour using Gaussian noise can introduce distribution differences. Additionally, the stitching of images may lead to discrepancies in distribution compared to real images. Chen et al.~\cite{10034609} demonstrated that splicing and cropping can create differences in image artifacts when compared to authentic images. These limitations highlight the need for further research and advancements in Deepfake generation to address issues related to realistic backgrounds, distribution differences, and counterfeit traces resulting from localized modifications and image operations.

Diffusion models~\cite{rombach2022high, ho2020denoising, dhariwal2021diffusion} have  received considerable attention recently due to their high quality visual content generation. These models also do not need to align the posterior distribution like VAE~\cite{kingma2019introduction}, or train  additional discriminators like GANs~\cite{goodfellow2020generative}. They aim  to transform data distribution into a tractable probability distribution, and then learn the reverse process in a step-by-step manner by parameterzing the reverse kernel with neural models. Though the generative abilities of diffusion models are remarkable, their sampling process has intrinsic randomness~\cite{cao2022survey}. This leads to a lack of control over the generated visual content. We find that generally, even detailed prompts fail to reasonably control the content of the generated images. This makes forgery of topic-oriented content with diffusion models  problematic. Moreover, the current models also lack in generating high-quality images of events involving interactions between multiple people. Our technique particularly focuses on these aspects. By guiding the image generation process with an anchor image, we provide consistency between multiple images that can collectively depict a forged event. We also improve the quality of the generated content that involves interaction between multiple people. 

\begin{figure}[t]
\centering
\includegraphics[ width= 0.5\textwidth]{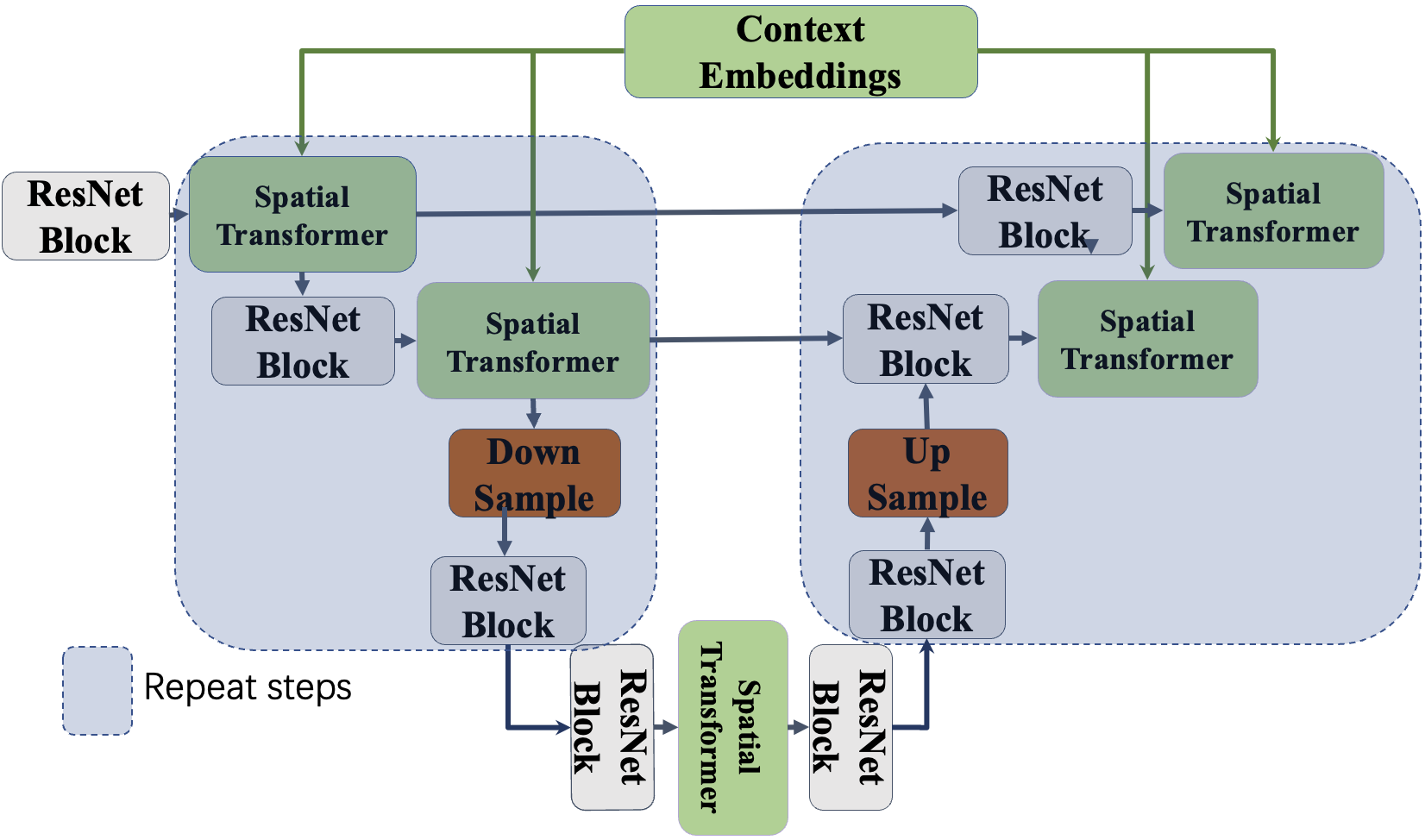}
\caption{The U-Net serves as the key component for generating images from noise. During the prediction process, the UNET is invoked multiple times, where each invocation removes the noise slice output by the previous UNET prediction from the original noise, gradually obtaining a denoised image representation. The UNET utilized in the SDM consists of approximately 860 million parameters and requires around 3.4 gigabytes of storage space when encoded with float32 precision.
}
\label{oldunet}
\end{figure}

In the past year, most popular diffusion models have utilized the convolutional U-Net~\cite{ronneberger2015u} as their backbone. The U-Net architecture involves changing the dimensions of latent variables through a combination of downsampling and upsampling operations, as illustrated in Fig.~\ref{oldunet}. The entire model can be understood as a sequence of denoising autoencoders with equal weights. To fully harness the potential of U-Net, we have modified its framework to accept both tokenized text and image inputs. To further improve the performance of the model, we have leveraged the fine-tuning capabilities of DreamBooth, a deep-learning generation model designed to enhance text-to-image models. By utilizing a real-person image as a reference, our objective is to train the default model to generate specialized individuals or objects. Furthermore, we have evaluated the stability of the image-to-image function within the diffusion process. This evaluation involves measuring the difference between the generated images and the original images using a denoising strength parameter. When the denoising strength is set to 0, the generated image will be an exact replica of the input image. However, as the denoising strength increases towards 1, the generated image will become increasingly distinct from the original image.

In summary, existing face forgery techniques suffer from blurring issues in both the generated face area and the non-face areas, particularly in the background of the image. Additionally, the randomness present in the generation process of current diffusion models does not align with the requirements of Deepfake generation tasks. The objective of our work is to address these challenges by combining the strengths of Deepfake generative models and Text-to-Image generative models, resulting in a novel image generation technique. Our approach aims to overcome the blurred area problem by introducing an input image that encompasses the entire scene. By considering the person and the background as a unified entity in the generative task, we effectively reduce counterfeit traces introduced through operations such as image cropping and splicing. Furthermore, this approach enhances the authenticity of the picture background. To enable the diffusion model to generate more specific and targeted images, we have modified the internal structure of the traditional U-Net. The new U-Net not only takes text as input but also incorporates a latent representation of a related image in the generation process. Moreover, unlike most diffusion models that focus on a single object during generation, our proposed model caters to scenarios involving multiple generated targets. It leverages a latent anchor image as guidance to improve the content of the generated images, thereby better meeting the requirements of Deepfake generation tasks.

\section{PROPOSED METHOD}

We propose Text-image guided Diffusion Model (TIDM), a novel technique aimed at generating realistic and consistent images for events such as meetings. The generation of high-quality images conditioned on text inputs remains a challenging task in the field of text-to-image synthesis. In this work, we address the limitations of existing methods, particularly the popular Stable Diffusion Model (SDM)~\cite{rombach2022high}, which is specifically designed for text-to-image generation tasks. SDM is based on diffusion process, which employs an iterative process of ``denoising" data in a latent representation space and subsequently decoding it to form complete images. However, while SDM has shown promising results, it still faces certain challenges, as discussed in Section~\ref{sec:RW}, which limit its ability to generate fake content for events like meetings. To overcome these challenges, we introduce the Text-image guided Diffusion Model (TIDM). Our proposed method leverages the strengths of SDM while incorporating additional improvements. In TIDM, we enhance the generation accuracy by introducing a new anchor image as a guiding reference in the U-Net architecture. This anchor image provides additional contextual information during the generation process, leading to improved image synthesis results.

We also develop a Dreambooth method based technique~\cite{ruiz2022dreambooth} for fine-tuning the TIDM model. Dreambooth enables targeted adjustments to the generated images, further enhancing their quality and coherence. By integrating dreambooth into TIDM, we improve the fine-grained details and overall realism of the generated images. To facilitate the generation process, TIDM utilizes the Contrastive Language-Image Pre-training (CLIP)\cite{luisier2010image} to convert input text into a text embedding. This text embedding is then fed into the denoising module known as the Text-conditioned Latent U-Net\cite{ronneberger2015u}. The Text-conditioned Latent U-Net processes the text embedding and outputs high-resolution images with dimensions of 512x512 pixels. Our proposed TIDM technique overcomes the limitations of SDM by incorporating the anchor image as a guiding reference and fine-tuning through the dreambooth method. These improvements lead to enhanced image synthesis performance, ensuring that the generated images accurately correspond to the input text descriptions.

\begin{figure*}[t]
\centering
\includegraphics[ width= 1\textwidth]{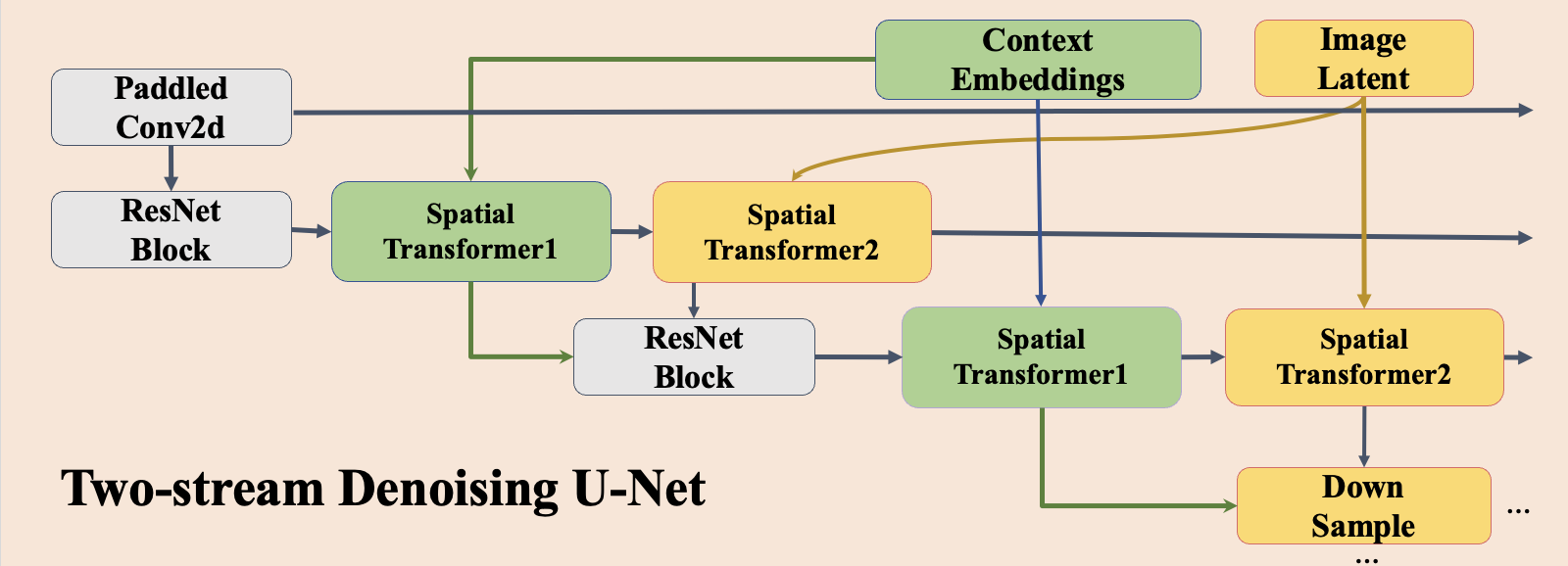}
\caption{TIDM is our enhanced denoising model that incorporates additional yellow blocks to handle both the image and text prompt representations in the Image Information Creator of SDM-based pipeline. By utilizing a two-stream architecture, TIDM effectively processes and integrates information from both modalities to generate denoised outputs. The inclusion of the yellow blocks enhances the model's capability to effectively denoise images while considering the accompanying text prompts, resulting in improved overall performance. 
}
\label{new unet.png}
\end{figure*}

TIDM introduces several modifications to address the limitations of SDM~\cite{rombach2022high} as a source for DeepFake generation, particularly when used to support rumors with fake evidence. The framework of TIDM is illustrated in Fig.~\ref{Inference process with new U-net architecture}. By incorporating an additional Image Encoder module, TIDM enables image guidance to the SDM backbone, resulting in improved image generation. The Image Encoder module in TIDM computes a latent representation for an anchor image, which serves as a visual reference to guide the generation of fake images. This latent representation is further processed by the Image Information Creator module, which also takes into account the text prompt for the fake content. Unlike SDM, our Image Information Creator has the unique ability to simultaneously process both the text prompt and the anchor image. To achieve this, we employ a two-stream denoising U-Net model, as depicted in Fig.~\ref{new unet.png}.  
Overall, the introduction of the anchor image and the utilization of the two-stream denoising U-Net model in TIDM significantly enhance the quality of generated images. These modifications enable TIDM to produce more convincing and realistic DeepFake images by incorporating relevant visual guidance during the generation process.

To mitigate the problems of blurred backgrounds and ensure a coherent representation of both the person and the background in the generative task, we adopt an approach that treats them as a unified entity. By considering the person and the background together, we reduce the likelihood of introducing artifacts that may arise from operations like cropping and splicing. This approach enhances the overall realism of the generated background, resulting in more visually convincing output. In Section~\ref{subsec:SDM}, we describe our improved denoising process for the SDM to enable our two-stream U-Net, while Section~\ref{sec:DB} explains how we utilized Dreambooth~\cite{ruiz2022dreambooth} to further improve the output quality.

\subsection{Objective Functions}
\label{subsec:SDM}

The overall learning objective of diffusion models can be expressed as minimization of the following loss~\cite{ho2020denoising}
\begin{equation}
L_{D M}=\mathbb{E}_{x, \epsilon \sim \mathcal{N}(0,1), t}\left\|\epsilon-\epsilon_\theta\left(x_t, t\right)\right\|_2^2 \label{1},
\end{equation}

where, $t$ is sampled from ${1,2, \dots, T}$; $x_t$ is a noisy image of size defined by the user, and $\epsilon$ and $\epsilon_\theta$ denote the standard normal distribution samples and denoising autoencoder outputs, respectively. Latent diffusion modeling operates in the latent space to reduce computational complexity of the learning objects, and employs the following loss.

\begin{equation}
L_{L D M} :=\mathbb{E}_{\mathcal{E}(x), \epsilon \sim \mathcal{N}(0,1), t}\left\|\epsilon-\epsilon_\theta\left(z_t, t\right)\right\|_2^2. 
\label{equ:SDM}
\end{equation}
In above, the latent representation $z_t$ can be efficiently obtained from the encoder $\mathcal{E}$ during training, and samples from $p(z)$ can be decoded to the image space with a single pass through a decoder $\mathcal{D}$.

Our network architecture features an input latent representation that also incorporates the latent of an anchor image from the start. This representation is computed by an image encoder, which is implemented in our technique using the Huggingface Diffusers library~\cite{githug}. Our image encoder uses  \texttt{down\_blocks}, \texttt{mid\_block}, and \texttt{up\_blocks}. 
We further enhance the encoder by incorporating a specialized PyTorch ModuleList that contains \texttt{CrossAttnDownBlock2D} and  \texttt{ResnetBlock}. TIDM's most important ability is to simultaneously process two-stream input of text and anchor image. To enable that, we must improve upon the denoising process of the SDM technique. We achieve that with a conditional denoising autoencoder, thereby effectively altering the loss function in Eq.~(\ref{equ:SDM}) to the following: 

\begin{equation}
L_{L D M} :=\mathbb{E}_{\mathcal{E}(x), \epsilon \sim \mathcal{N}(0,1), t}\left\|\epsilon-\epsilon_\theta^*\left(z_t, t\right)\right\|_2^2,
\label{equ:new unet}
\end{equation} 
where $\epsilon_\theta^*$ denotes the conditional denoising autoencoder output. 
Our U-Net based architecture for the denoising process is shown in Fig.~\ref{new unet.png}. Note that, this architecture is an enhancement of the existing architecture that is known to work well for the diffusion model~\cite{rombach2022high}, but was previously incapable of processing both text and visual input.

In the original Stable Diffusion Model (SDM), the user input is limited to text content and inference hyper-parameters, which determines the content of the generated images. However, this loose control over the output often leads to inconsistent and visually inferior results. To address these limitations, we propose the Text-image guided Diffusion Model (TIDM), which introduces a control mechanism using an anchor image for significant improvements. Generating realistic faces and body part shapes becomes challenging for traditional diffusion models, such as SDM, when the text prompts involve multiple personalities. 
To address this limitation, our approach overcomes the challenge by incorporating an anchor image. This incorporation enables us to have more precise control over the generated images, enhancing their quality and realism. To achieve this, we easily access celebrity images from the Internet and utilize them as anchors in our generation process. By incorporating the anchor image, TIDM ensures that the generated images exhibit realistic face and body shapes, even when the text prompts involve multiple personalities. 

In addition to controlling the background, TIDM also focuses on generating high-quality foreground personnel. To achieve this, we employ a modified Dreambooth training method. Dreambooth allows targeted adjustments to the generated images, enhancing their quality and coherence. By integrating this method into TIDM, we are able to generate high-quality figures of the foreground personnel, ensuring visually appealing and convincing results. The combination of the anchor image control mechanism and the modified Dreambooth training method enables TIDM to generate images with improved consistency, visual quality, and realism. The use of an anchor image provides precise guidance for the generation process, enhancing the coherence between multiple output samples. Furthermore, the modified Dreambooth training method fine-tunes the generated images, resulting in visually appealing foreground personnel. By addressing the limitations of traditional diffusion models and incorporating control mechanisms with anchor images and modified training methods, TIDM offers a significant improvement in generating realistic and visually consistent DeepFake images. The ability to control both the background and foreground personnel enhances the quality and believability of the generated images, making them more suitable for various applications in the realm of DeepFake synthesis.

\subsection{Dreambooth Training Method}
\label{sec:DB}

Our objective is to generate detailed and diverse images of a subject using only a limited number of their casually obtained photos (3-5). To that end, the first task is to implant the subject instance into the output domain of the model and to bind the subject with a unique identifier. The identifier can be varied  depending on different objects users are interested in. We present our method to design the identifier below, as well as our approach to supervising the fine-tuning process of the model such that it can re-use its prior learning and adapt it to subject instance.
The Prior-Preservation Loss used by our technique under the Dreambooth framework is given below.
\begin{equation}
\begin{split}
\mathbb{E}_{\mathbf{x}, \mathbf{c}, \boldsymbol{\epsilon}, \boldsymbol{\epsilon}^{\prime}, t}
[w_t\left\|\hat{\mathbf{x}}_\theta\left(\alpha_t \mathbf{x}+\sigma_t \boldsymbol{\epsilon}, \mathbf{c}\right)-\mathbf{x}\right\|_2^2 
+\\
\lambda 
w_{t^{\prime}}\left\|\hat{\mathbf{x}}_\theta\left(\alpha_{t^{\prime}} \mathbf{x}_{\mathrm{pr}}+\sigma_{t^{\prime}} \epsilon^{\prime}, 
\mathbf{c}_{\mathrm{pr}}\right)-\mathbf{x}_{\mathrm{pr}}\right\|_2^2].
\label{2}
\end{split}
\end{equation}

Here, $\lambda$ controls the prior-preservation term weight, $x$ represents the ground-truth image, $c$ is a conditioning vector obtained from a text prompt, $\epsilon$ is Gaussian noise distribution, and $\alpha_t, \sigma_t, w_t$ are hyperparameters to regulate the noise schedule and sample quality. To match our model's input, we adapted the Dreambooth training process for output size.

The task of Dreambooth is to obtain a unique model by training the base model. The base model is obtained by stable diffusion, which lacks the ability to generate specific centers. We manipulated the structure of the denoising U-Net and tuned the input and output architecture parameters. The new architecture can obtain image input information to guide the generative model with stronger detail processing ability. For original Dreambooth, the input would be placeholder token and images used for training. The trained model include .ckpt file and related components such as a scheduler, text encoder, tokenizer, U-Net and VAE. We replaced the conventional denoising U-Net in the training files with our modified U-Net. Due to the different architectures, the dimensions of the output parameters were also modified.

\begin{figure}[t]
\centering
\includegraphics[ width= 0.5\textwidth]{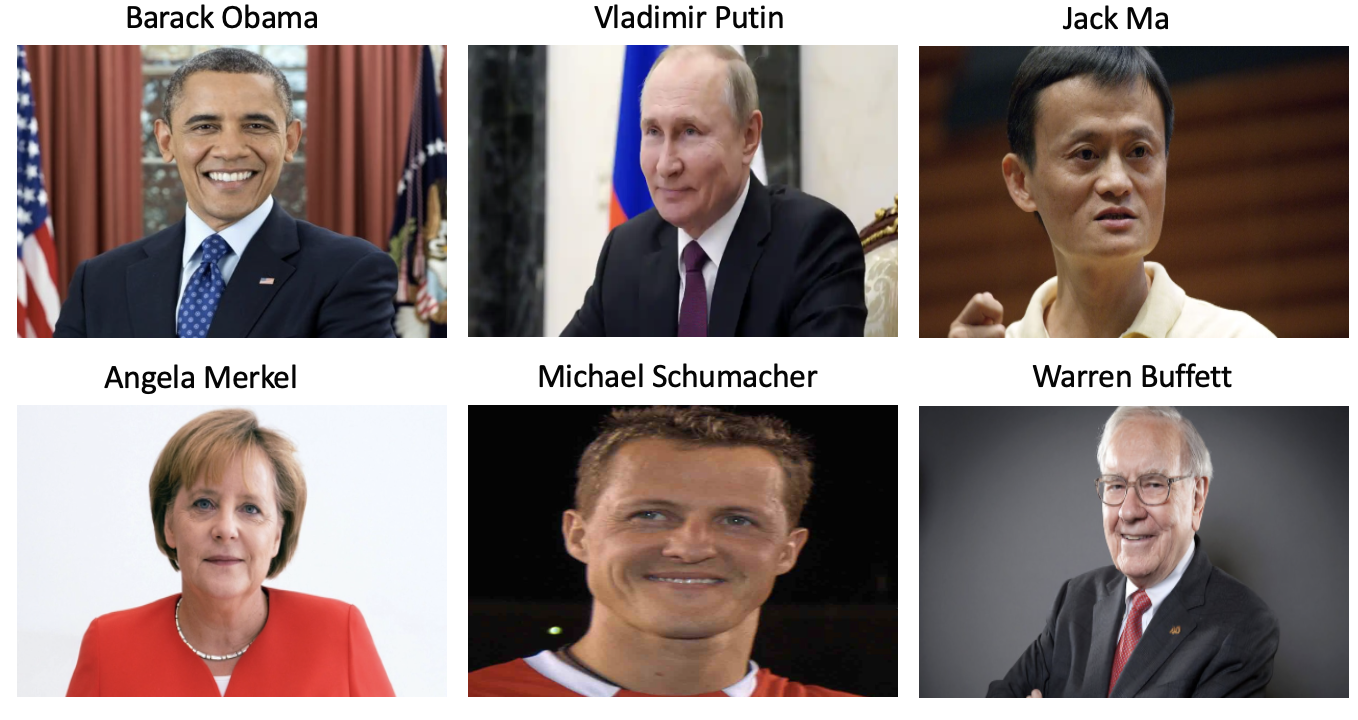}

\caption{Examples of anchor images obtained from a Google search. The name of each celebrity is located above the image.}
\label{Anchor}

\end{figure}

\begin{figure*}[t]
\centering
\includegraphics[ width= 1\textwidth]{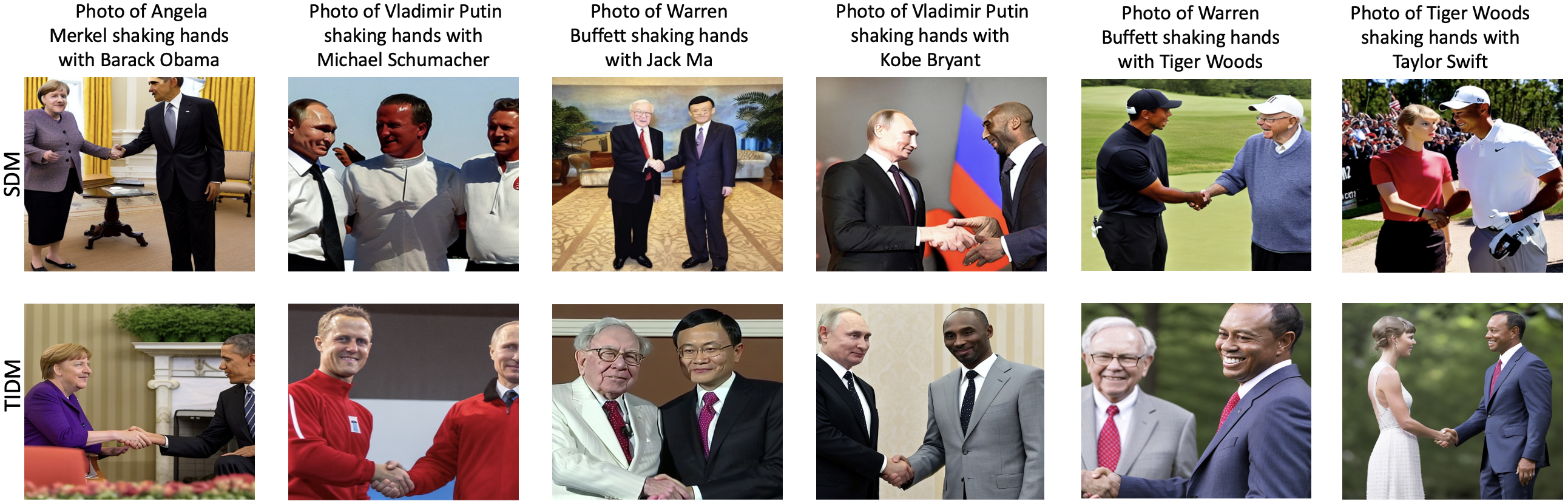}
\caption{
Comparison between SDM~\cite{rombach2022high} and TIDM generated outputs, using the corresponding text prompts noted at top. 
}
\label{SDM xi}
\end{figure*}

\section{EXPERIMENTS}
\label{sec:experiment}

We extensively evaluate the effectiveness of our method through a series of experiments. Our experiments specifically focus on generating forged images that depict meetings between celebrities. The subjects in our forged images primarily include influential leaders of different countries and celebrities with international recognition, such as South African President Cyril Ramaphosa and China's wealthiest individual Jack Ma. On one hand, the presence of prominent individuals in counterfeit visual content increases the likelihood of spreading misinformation. Conversely, fake content that includes well-known celebrities is relatively easy to evaluate subjectively, and obtaining high-definition images of these individuals is also more accessible. This reduced difficulty in counterfeiting makes such content a practical test-bed for evaluating the effectiveness of our technique.

\subsection{Datasets and Experimental Setup} 

Training images of the celebrities were obtained from Google searches. The selection of such images aimed to emulate a realistic scenario of generating counterfeit news through the manipulation of generative models. Specifically, we trained our private custom model using a limited number of pictures for each individual, ranging from $3$ to $5$, with the help of a modified Dreambooth training file. The training process was executed on a computer equipped with a 3090 graphics card, an i9-10900 CPU, and 128 GB of running memory. The total training time was less than 20 minutes. In Fig.~\ref{Anchor}, we have provided some examples of anchor images.

In our experiments, the images are resized to $512 \times 512$ pixels. The weights of the Huggingface model, which adopts the U-Net architecture and comprises 860 million parameters, are stored in a .bin file. To ensure fairness in comparisons, we adopt the Denoising Diffusion Implicit Model (DDIM)~\cite{song2020denoising} sampling method with 50 inference steps for image generation. The number of inference steps is typically positively correlated with the quality of generated images but also with the computational time. By employing the DDIM, we achieve a balance between image quality and computational time. The guidance scale for our model is 7.5, and to improve the utilization of computational resources, our model generates four images simultaneously. The inference process consists of two steps: (1) obtaining and saving the Dreambooth model locally or in the cloud; (2) importing the specific model into our custom pipeline to generate images. For successful image generation, the prompt must match the placeholder token taught in the Dreambooth model.

\subsection{Comparison with Stable Diffusion}

\begin{figure}[t]
\centering
\includegraphics[ width= 0.48\textwidth]{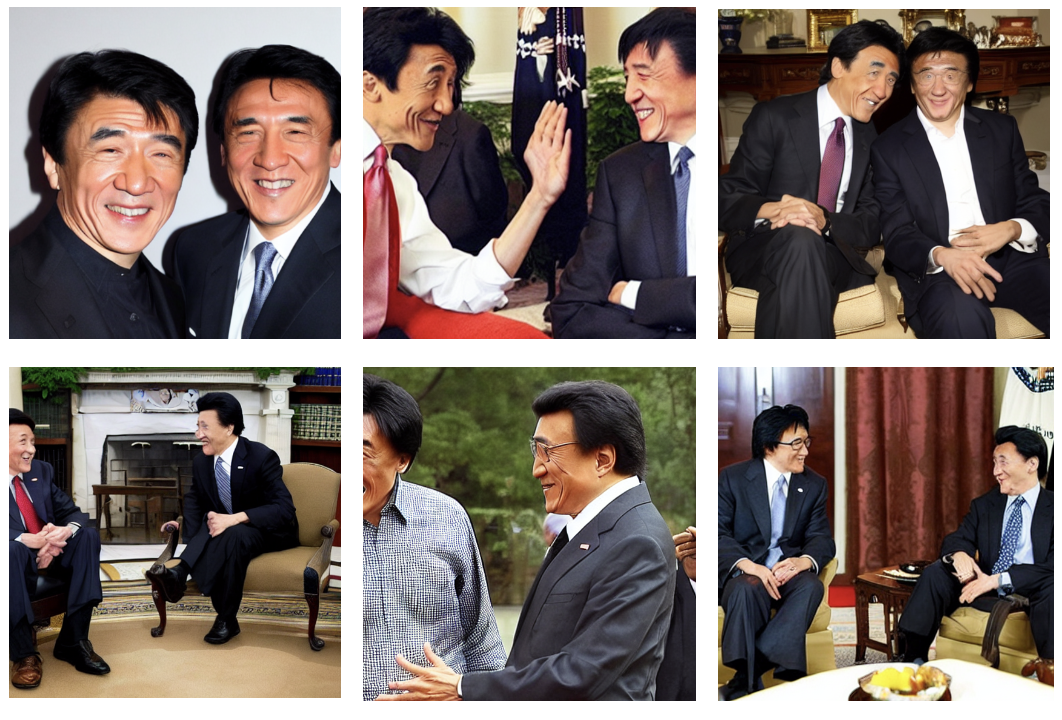}

\caption{ Some examples of facial fusion features obtained from the SDM when generating images based on the text prompt "a picture of Obama chatting with Jackie Chan"
}
\label{OX}
\end{figure}

Typical images generated by the SDM~\cite{rombach2022high} and the corresponding images of the proposed TIDM are displayed in  Fig.~\ref{SDM xi}.
Each image pair in the figure uses the same text prompt, mentioned at the top of each pair. In general, our inspection of the images generated by  SDM revealed that the realism of human faces and the background is not satisfactory. 
One notable challenge encountered by the SDM method is the blending of facial features when the input prompt involves multiple subjects. As depicted in Fig.~\ref{OX}, we utilize the text ``a picture of Obama chatting with Jackie Chan" as an input prompt and simultaneously generate 10 images. Among them, six images exhibit issues with facial feature fusion, where the faces of the two individuals in the picture appear quite similar. It is evident that the facial features primarily resemble Jackie Chan's face while also incorporating some of Obama's facial features. This phenomenon arises due to the limited ability of SDM to generate multiple centers. Consequently, most SDM examples feature only a single central object.

\begin{figure}[t]
\centering
\includegraphics[ width= 0.4\textwidth]{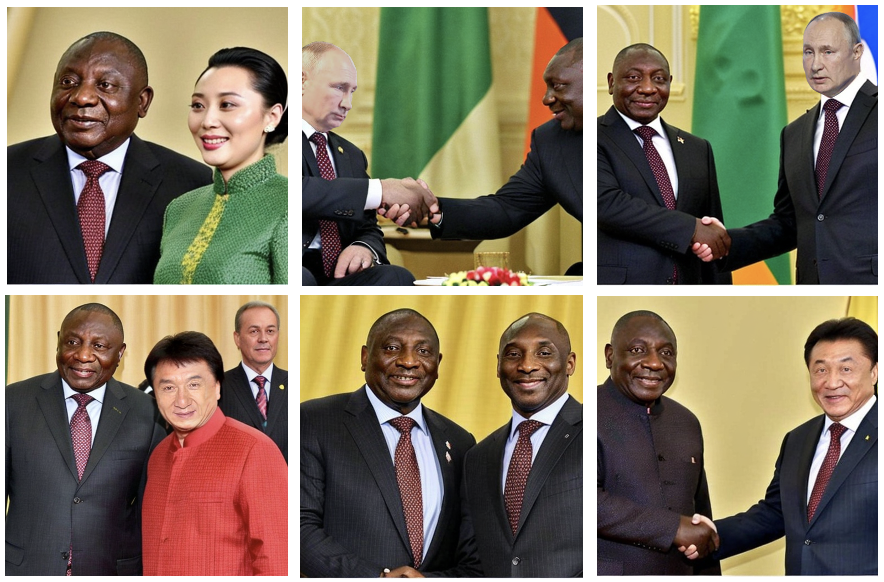}
\vspace{-3mm}
\caption{Images generated with TIDM by controlling the background with anchor images for consistency between  images.}
\label{Random}
\end{figure}

To address this problem, our model trained with the Dreambooth method, is able to use   a specific subject identifier. This enables us to  use  latent representation of a person's image as a secondary input. Our solution effectively resolves the blending problem. In Fig.~\ref{SDM xi}, the second row showcases the image output generated by TIDM. It is evident that our model has achieved a significantly higher level of image generation quality by training with the Dreambooth algorithm. 
our method clearly demonstrates a significant improvement in the image quality as compared to the SDM output. It is emphasized that the displayed images for the SDM are hand-picked from the visually appealing outputs. Mostly, the SDM images suffered from obvious problems, e.g., blending of faces, which are resolved by TIDM. In Fig.~\ref{Random}, we exemplify the background control with our method where multiple images of different celebrity meetings are generated with a background that can provide consistency between the images. This background control is possible with the proposed use of anchor image in the generation process.

\subsection{Subjective assessments}

 Subjective human evaluation of our model, compared to SDM model, was conducted with ten academics from various fields. We provided prompts for ten unique characters and generated ten images per prompt using both models. The participants rated the realism of the images on a scale of 1 (``obvious fake") to 10 (``very convincing"). The results of the evaluation are presented in Table~\ref{tab:2}, which shows that our model significantly outperformed the stable diffusion model in generating convincing images.

\begin{tiny}
\begin{table}[t]
\scriptsize
\centering
\renewcommand\arraystretch{1.1}
\setlength{\tabcolsep}{6mm}{
\caption{
Subjective evaluation of images with average rating from 10 evaluators using a 1-10 score range for convincingness across 10 instances of each corresponding Image Id.}

\label{tab:2}
\setlength\tabcolsep{2.8pt}
\begin{tabular}{p{0.9cm} c c c c c c c c c c | c}
\hline
Img. ID & 1 & 2 & 3 & 4 & 5 & 6 & 7 & 8 & 9 & 10 & Avg.\\
\hline
SDM & 7.10 & 6.00 & 7.00 & 7.30 & 6.38 & 7.00 & 6.40 & 7.80 & 8.60 & 6.00 & 6.96\\
TIDM & 9.80 & 8.60 & 7.90 & 9.20 & 8.80 & 8.60 & 8.44 & 8.10 & 8.80 & 8.20 & \textbf{8.64}\\
\hline
\end{tabular}
}

\end{table}
\end{tiny}

\section{Conclusion}

This paper introduces an advanced method for generating highly realistic fake images that can be utilized as convincing evidence of events, such as meetings involving prominent public figures. Additionally, the proposed approach enables the generation of multiple images with consistent backgrounds to support rumors or misleading narratives. The primary objective of this research is to emphasize the potential risks associated with AI-generated visual content in the dissemination of fake news. By raising awareness about these dangers, we aim to encourage proactive measures to mitigate the negative impacts of misleading and manipulated visual media.

\section{ACKNOWLEDGMENTS}

This research was supported by National Intelligence and Security Discovery Research Grants  (project$\#$ NS220100007), funded by the Department of Defence Australia.

\bibliographystyle{plain}
\bibliography{bibliography.bib}

\end{document}